# Consistency of UML class, object and statechart diagrams using ontology reasoners ☆

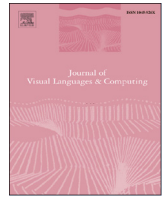

## Ali Hanzala Khan *, Ivan Porres

*Department of Information Technologies, Åbo Akademi University Joukahaisenkatu 3-5, FI-20520 Turku, Finland*

A B S T R A C T

We propose an automatic approach to analyze the consistency and satisfiability of Unified Modeling Language UML models containing multiple class, object and statechart diagrams using logic reasoners for the Web Ontology Language OWL 2. We describe how to translate UML models in OWL 2 and we present a tool chain implementing this translation that can be used with any standard compliant UML modeling tool. The proposed approach is limited in scope, but is fully automatic and does not require any expertise about OWL 2 and its reasoners from the designer.

© 2014 The Authors. Published by Elsevier Ltd. This is an open access article under the CC BY license (http://creativecommons.org/licenses/by/3.0/).

## 1. Introduction

Model Driven Engineering (MDE) [1] advocates the use of models to represent the most relevant design decisions of a software development project. A MDE software development project involves the creation of many models. Each model is used for describing, visualizing and observing different viewpoints of a system at different levels of abstractions [2].

A software model usually comprises a number of diagrams. The diagrams in a software model are described using a particular modeling language. A well-known general modeling language used by practitioners during software development process is the Unified Modeling Language (UML) [2,3]. The definition of a modeling language is given in terms of a metamodel by using a metamodeling language, such as Meta Object Facility (MOF) [4] or Kernel Meta Meta Model (KM3) [5]. This paper focuses on the analysis of models specified using UML superstructure specification [2] and MOF.

UML models can be represented in the form of a theory in mathematical logics [6], such as, description logics [7] or predicate logics [8]. A *consistent* logical theory is the one which does not contain a contradiction or an unsatisfiable concept [9,10]. Similarly, we consider a model to be consistent if it does not contain an unsatisfiable concept.

The presence of concepts in a model that are not satisfiable reveals design errors. For example if a UML class diagram depicts unsatisfiable classes, then it is not possible to instantiate objects conforming to theses classes. Furthermore, in case of an inconsistent behavioral diagram (such as inconsistent statechart diagrams), an object cannot enter into an unsatisfiable state.

The unsatisfiable concepts in models should be identified as early in the development process as possible. In this paper we propose an approach that automatically checks the consistency and satisfiability of UML models. If a model is found to be inconsistent, then the proposed approach indicates the unsatisfiable concepts that make the whole model inconsistent.

We call the task of finding out the inconsistencies in software models a *model validation*. The validation of modeling artifacts has been discussed in many research papers. However, we consider that there is still need for more

☆ This paper has been recommended for acceptance by Shi Kho Chang.
* Corresponding author. Tel.: +358 2 215 3463.
*E-mail addresses:* ali.khan@abo.fi (A.H. Khan),
ivan.porres@abo.fi (I. Porres).

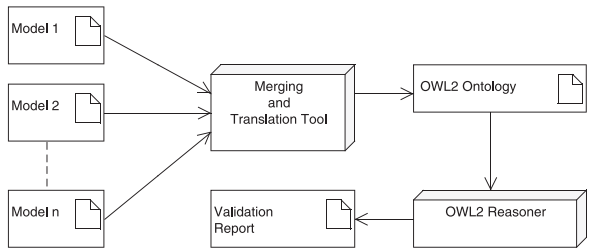

**Fig. 1.** Workflow of the proposed consistency checking approach.

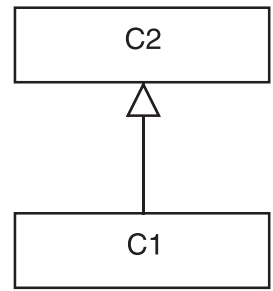

**Fig. 2.** A UML model depicting a class C1 being a subclass of a class C2.

research in this area due to the vast number of different validation problems that exist for complex models.

Among the validation problems in behavioral diagrams that has already been discussed by other researchers there are for instance, analysis of the control looping to find deadlocks [11], analysis of method invocations against the class description for finding deadlocks [12]. Also, checking the consistency of statechart diagrams and class diagrams by using the $\pi$-calculus [13]. The research on validation problems of structural diagrams is also very vast. A number of problems that have been discussed in the recent past by other researchers include the consistency of UML class diagrams with hierarchy constraints [14], the reasoning of UML class diagrams [15], the full satisfiability of UML class diagrams [16], and the inconsistency management in model driven engineering [17]. Although, a lot of research work has already been done in the area of the validation of structural and behavioral diagrams, we still believe there is a room for new approaches in this area.

The validation problem that we tackle in this paper can be stated as follows: Is a model containing multiple UML class diagrams, UML object diagrams and UML statecharts containing class and state invariants consistent and satisfiable? Model consistency and satisfiability is established by translating the models into a logical theory, and then using automatic logical reasoners to infer the logical consequences of the translated models. More concretely, we propose to represent UML models using a description logic by means of the OWL 2 DL language [18,19], and to analyze the translated models using automated OWL 2 reasoners [20,21]. The approach we present in this paper is fully automated and it is implemented in the form of a tool that can be used with any standard compliant UML modeling tool.

In order to implement a fully automatic tool, we have decided to use description logic as the underlying formalism for our approach and OWL 2 DL [19] as the language to represent the UML models internally. This decision is supported by the fact that there are reasoners for description logic with the efficient decision procedures that are automatic [20,21]. Alternatively, there is a number of theorem proving tools available that are based on a high order logic, such as HOL [22]. These tools are very powerful but they require interaction with an expert human user. Also, there are model finders, such as Alloy [23] or Microsoft formula [24], which are automatic, but require that we artificially limit the search space.

The workflow of our approach is shown in Fig. 1. A number of UML models are taken as an input. All the inputs are translated to a decidable fragment of OWL 2, i.e.,

OWL 2 DL [19]. In the next step, the OWL 2 translations of UML diagrams are passed to a reasoner in the form of an ontology. The reasoner processes the ontology and produces a validation report. The validation report reveals the inconsistencies in the ontology representing the UML models. The detailed discussion about the contents of the validation report is discussed later in different sections of this paper.

In this paper we address the issues that have been inadequately or not addressed in the previous research. The novelty of our work is that we offer the validation of many modeling concepts under one approach. The modeling concepts that can be validated using our approach include the following: classes, objects, associations, links, labeled links, domain and range, multiplicity, composition (herein unshearedness and acyclicity), unique and non-unique associations, ordering, class generalization, and association generalization. Furthermore, the proposed approach also allows us to analyze the conformance of object diagrams against class diagrams, consistency of class diagrams and statchart diagrams, consistency of state invariants written using a subset of object constraint language, and the consistency of multiple models when merged together.

This paper also contains several applications of the proposed approach. These include (1) validation of multiple models of the same metamodel when merged, (2) validation of class and object diagrams with OCL invariants and (3) validation of class and statechart diagrams with OCL state invariants. The detection of errors in the above-mentioned models, and the results of the performance tests of the proposed approach that is shown in this paper is the evidence that the proposed approach is viable and practical, and can be applied in the industry.

In the next section we give an overview of this research.

## 2. Background

### 2.1. Ontology foundations

An ontology is a specification of a conceptualization [25]. In this paper our understanding of the term "*specification of a conceptualization*" is the specification of concepts and relationships that can represent an abstraction of a program. The abstraction of a program is typically expressed in the form of models by using modeling languages. For example the fact that a class C1 is a subclass of a class C2 is drawn by using UML, as given in Fig. 2.

In logics the abstraction of a program is expressed in the form of logical facts using logical languages such as description logic [18] or predicate logic [8]. For example the fact depicted in Fig. 2 that the class C1 is a subclass of the class C2 is written in description logics as

$$C1 \sqsubseteq C2$$

In ontologies the concepts are represented in the form of axioms that depict the specification of a conceptualization [25]. These axioms represent concepts as classes, and the relationship among concepts as properties. Since the ontology deals with concepts and their relationships, the language used for writing an ontology is semantically very close to the language used to express the logic [25]. This allows us to write the logical facts as axioms in the ontology. For example, the specialization relation $C1 \sqsubseteq C2$ (shown in Fig. 2) is written in ontology as

```
SubClassOf(C1 C2 )
```

In this paper we use ontologies to represent the semantics of UML models as logical facts, so that we may able to infer the logical consequences from these logical facts automatically using a reasoner.

### 2.2. Description logic

The description logic (DL) used in our approach is classified as $\mathcal{SROIQ}$ [18]. Description logic is made up of concepts, denoted here by $C, D$, and roles, denoted here by $R, Q$. A concept or role can be named, in which case it is called atomic, or it can be composed from other concepts and roles.

An interpretation $\mathcal{I}$ consists of a non-empty set $\Delta^{\mathcal{I}}$ and an interpretation function which assigns a set $C^{\mathcal{I}} \subseteq \Delta^{\mathcal{I}}$ to every named concept $C$ and a binary relation $R^{\mathcal{I}} \subseteq \Delta^{\mathcal{I}} \times \Delta^{\mathcal{I}}$ to every named role $R$.

The constructors of description logic are as follows:

| Everything | $\top^{\mathcal{I}}$ | $=$ | $\Delta^{\mathcal{I}}$ |
|---|---|---|---|
| Nothing | $\bot^{\mathcal{I}}$ | $=$ | $\varnothing$ |
| Complement | $(\neg C)^{\mathcal{I}}$ | $=$ | $\Delta^{\mathcal{I}}/C^{\mathcal{I}}$ |
| Inverse | $(R^-)^{\mathcal{I}}$ | $=$ | $\{(y, x) \| (x, y) \in R^{\mathcal{I}}\}$ |
| Intersection | $(C \sqcap D)^{\mathcal{I}}$ | $=$ | $C^{\mathcal{I}} \cap D^{\mathcal{I}}$ |
| Union | $(C \sqcup D)^{\mathcal{I}}$ | $=$ | $C^{\mathcal{I}} \cup D^{\mathcal{I}}$ |
| Restriction | | | |
| Universal | $(\forall R.C)^{\mathcal{I}}$ | $=$ | $\{x \| \forall y.(x, y) \in R^{\mathcal{I}} \to y \in C^{\mathcal{I}}\}$ |
| Existential | $(\exists R.C)^{\mathcal{I}}$ | $=$ | $\{x \| \exists y.(x, y) \in R^{\mathcal{I}} \wedge y \in C^{\mathcal{I}}\}$ |
| Cardinality | $(\geq nR)^{\mathcal{I}}$ | $=$ | $\{x \| \#\{y \| (x, y) \in R^{\mathcal{I}}\} \geq n\}$ |
| | $(\leq nR)^{\mathcal{I}}$ | $=$ | $\{x \| \#\{y \| (x, y) \in R^{\mathcal{I}}\} \leq n\}$ |

where $\#X$ is the cardinality of $X$. The axioms in DL can be either inclusions $C \sqsubseteq D$, $C \sqsubseteq D$ or equalities $C \equiv D$, $R \equiv Q$.

An interpretation satisfies an inclusion $C \sqsubseteq D$ if $C^{\mathcal{I}} \subseteq D^{\mathcal{I}}$ and an inclusion $R \sqsubseteq Q$ if $R^{\mathcal{I}} \subseteq Q^{\mathcal{I}}$. An interpretation satisfies an equality $C \equiv D$ if $C^{\mathcal{I}} = D^{\mathcal{I}}$ and an equality $R \equiv Q$ if $R^{\mathcal{I}} = Q^{\mathcal{I}}$. $\mathcal{I}$ satisfies a set of axioms if it satisfies each axiom individually – $\mathcal{I}$ is then said to be a model of the set of axioms. Given a set of axioms $\mathcal{K}$, a named concept $C$ is said to be satisfiable if there exists at least one model $\mathcal{I}$ of

$\mathcal{K}$ in which $C^{\mathcal{I}} \neq \varnothing$. A set of axioms is said to be satisfiable if all of the named concepts that appear in the set are satisfiable. If a set of axioms $\mathcal{K}$ is satisfiable, we say that an axiom $\phi$ is satisfiable with respect to $\mathcal{K}$ if $\mathcal{K} \cup \{\phi\}$ is satisfiable. Similarly, we say that $\phi$ is unsatisfiable (w.r.t. $\mathcal{K}$) if $\mathcal{K} \cup \{\phi\}$ is unsatisfiable.

The decidability of $\mathcal{SROIQ}$ is demonstrated by Horrocks et al. [18], and there exist several reasoners that can process answer satisfiability problems automatically [20,21].

### 2.3. Web Ontology Language OWL 2

The Web Ontology language OWL 2 [26] is a language for defining ontologies. The OWL 2 provides axioms to express model-theoretic semantics which are compatible with the DL $\mathcal{SROIQ}$ such as classes, properties, individuals and data values [26]. The OWL 2 is also supported by logic reasoners such as Pellet [20] and HermiT [21]. In this paper, we use the OWL 2 functional syntax (OWL2fs) [19] to explain the translations of UML concepts to OWL 2 axioms. The UML to OWL 2 translation that we propose in this paper is also implemented in the form of a translation tool. The translation tool generates OWL 2 files that contain the translations of UML models in two different syntaxes such as OWL2fs and OWL 2 Manchester syntax. The reason of generating outputs in two different OWL 2 syntaxes is because we want to analyze the performance of different reasoners. The detail about the performance of different reasoners will be discussed later in this paper. The interpretation of the main OWL 2 expressions used in this paper is shown in Table 1. A complete description of the OWL 2 semantics, including support for data types can be found in [26].

**Example.** The OWL 2 translation of the UML model shown in Fig. 3 is as follows:

```
Declaration(Class(C1))
Declaration(Class(C2))
Declaration(ObjectProperty(A))
ObjectPropertyDomain(A C1 )
ObjectPropertyRange(A C2 )
SubClassOf(C1 ObjectMinCardinality(min A ) )
SubClassOf(C1 ObjectMaxCardinality(max A ) )
```

**Table 1**
DL interpretation of the main OWL 2 expressions used in this paper.

| OWL 2 | DL |
|---|---|
| SubClassOf(C D) | $C \sqsubseteq D$ |
| EquivalentClasses(C D) | $C \equiv D$ |
| DisjointClasses(C D) | $C \sqcap D = \varnothing$ |
| ObjectIntersectionOf(C D) | $C \sqcap D$ |
| ObjectUnionOf(C D) | $C \sqcup D$ |
| SubObjectPropertyOf(R1 R2) | $R1 \sqsubseteq R2$ |
| Inv.ObjectProperties(R1 R2) | $R1 \equiv R2^-$ |
| Inv.FunctionalObjectProperty(R) | $\top \sqsubseteq (\leq 1R^-)$ |
| ObjectPropertyDomain(R C) | $\exists R.\top \sqsubseteq C$ |
| ObjectPropertyRange(R C) | $\top \sqsubseteq \forall R.C$ |
| ObjectMinCardinality(n R) | $\geq nR$ |
| ObjectMaxCardinality(n R) | $\leq nR$ |
| ObjectExactCardinality(n R) | $(\geq nR) \sqcap (\leq nR)$ |
| ClassAssertion(C x) | $C(x)$ |
| ObjectPropertyAssertion(R x y) | $R(x, y)$ |
| Neg.Obj.PropertyAssertion(R x y) | $\neg R(x, y)$ |

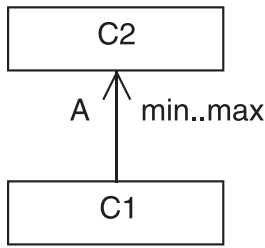

**Fig. 3.** The UML Model representing two classes C1 and C2 connected with each other using the association A.

## 2.4. OWL and UML

The Object Management Group (OMG) [27] specifies the Ontology Definition Metamodel (ODM) [28] that makes the concepts of OMG modeling hierarchy applicable to the ontology engineering [28]. The ODM follows the similar hierarchy as the one mentioned in a four layer OMG modeling hierarchy.

The ODM is a specification of ontology structure, and it is derived from MOF [28,29]. It comprises of classes, associations and constraints [29, Chapter 7].

The core of ODM represents formal logic languages, such as description logic (DL), common logic (CL) and first-order predicate logic. The scope of these languages covers the representations of higher order complex representations of simple taxonomic expressions [28].

The other metamodel derived from MOF that is essential for ODM is the UML [28,29]. The UML is the widely used modeling language for the designing of conceptual and logical models. The UML is also derived from MOF, and there exist commonalities between UML and ODM specification. Therefore UML notations are also used for ontology modeling [29, Chapter 7].

The UML includes a number of metamodels that provide the modeling specification of languages such as OWL and Resource Definition Framework [30]. In our approach, we use ODM metamodel OWL in order to represent the MOF/UML based models. In our approach we use a decidable fragment of OWL 2 that is based on DL. This fragment is known as OWL 2 DL.

The UML and ODM are derived from MOF [28,29], therefore there exist commonalties, as well as differences between them. The summary of the common features of UML and ODM is shown in Table 2. Since in this research we use OWL 2 DL as a specification language for ontologies, the comparison between UML and OWL 2 DL mentioned here is given in terms of UML and OWL 2 DL.

There are some features in UML which do not have equivalent OWL 2 elements (see Table 3). In such cases we translate the UML elements by using a combination of existing OWL elements which have an equivalent DL meaning as UML. These translations will be discussed later in this paper in different sections.

## 2.5. Reasoners

A reasoner is a utility that automatically infers the logical consequences from a set of logical facts [31]. We have several selection criteria for the reasoners, which include, first, a complete support for OWL 2 and Semantic Web Rule Language (SWRL, discussed later in this paper), and second, the reasoner is freely available as open source. The first requirement is motivated by the ontologies used in our research. The second requirement ensures that the research is easily repeatable by others.

**Table 2**
UML elements which have the DL equivalent OWL 2 DL elements.

| UML elements | OWL elements |
| --- | --- |
| Class | Class |
| Instance | Individual |
| Binary association | Property |
| Binary association link | Property assertion |
| Class generalization | Subclass |
| Property generalization | Subproperty |
| Enumeration | Oneof |
| Multiplicity | Min/max/exact cardinality |
| Navigation | Domain/range |
| Datatype | Datatype |

**Table 3**
UML elements which do not have the DL equivalent OWL 2 DL elements.

| UML elements | OWL elements |
| --- | --- |
| Ordering | Not available |
| Composition | Not available |
| Composition unshearedness | Not available |
| Composition acyclicity | Not available |
| Non-unique properties | Not available |
| Label on a link | Not available |
| State | Not available |
| Transition | Not available |
| State invariant | Not available |
| OCL constructs | Not available |

Based on these criteria, we have chosen the following two reasoners:

1. Pellet [20]: An open source Java-based ontology reasoner developed by Clark and Parsia LLC, which is an R&D firm, specializing in Semantic Web and advanced systems.
2. HermiT [21]: An open source reasoner that is implemented in Java, and developed by the Information Systems Group of Oxford University.

Both reasoners are implemented in Java, offer complete support for OWL 2 and SWRL and are freely available as open source, which satisfy our requirements.

## 3. Related work on model validation

In this section, we discuss the most important related works done by other researchers in the area of model validation. The discussion included in this section is categorized according to the research objectives of this research.

## 3.1. Consistency of class and statechart diagrams

The use of ontology languages and description logic in the context of model validation has been proposed in the past by different authors [17,32,15,33]. However, to the best of our knowledge, none of them has addressed the reasoning of the satisfiability of state invariants using OWL 2 DL. These works focus on the problem of class diagrams satisfiability, i.e. whether a class diagram can generate consistent object diagrams or not. Furthermore, the TWOUSE approach [34] is focused on two areas: the first is the ontology development modeling, and the second is the translation and validation of Domain Specific Languages (DSL) by using OWL 2. The TWOUSE approach proposed the same methodology for the validation of DSLs as presented in this paper. However, their work on validation is limited to the validation of DSLs, and has not yet offered the validation of statechart diagrams with state invariants.

Yeung [11] analyzed control looping to find deadlocks, by translating class diagrams into the B-Method and statechart diagrams into CSP. Their approach does not focus on the consistency of state invariants, the translation is done manually and there is no discussion about the verification method, whether it is manual or automatic.

Rasch et al. [12] used Object-Z for the formalization of class diagrams and CSP for statechart diagrams. Their approach analyzes method invocations against the class description and finds deadlocks by running the class and statechart diagram formalization in FDR. Their approach is not focused on analyzing the consistency of state invariants.

Vitus and Lam [13] analyzed the consistency of state-chart diagrams and class diagrams by using the $\pi$-calculus. The translation of UML diagrams to $\pi$-calculus is done manually and the consistency is analyzed by running the $\pi$-calculus script on the Workbench. Their work is not analyzing the consistency of state invariants.

Emil Sekerinski [35] verified the UML statecharts, in which the events are manually translated into generalized program statements, and these statements appear as the body of a transition. The execution of the program statements is based on the assumption that the body of the transition can read or write the values of the class variables.

Moaz et al. [36] analyzed the consistency of class and object diagrams by using Alloy. It is a fully automatic approach, in which the class and object diagrams are first translated into a parameterized Alloy module, and then the consistency analysis is done by analyzing the translated Alloy module by using the Alloy Analyzer. Their approach does not yet support statecharts and OCL constraints.

The state invariants are usually written by using the Object Constraint Language (OCL). It is the widely accepted language for writing constraints over UML diagrams. The reason why the existing research is not focused on analyzing the satisfiability of state invariants is because, in general, OCL is undecidable. However, the undecidability can be avoided, if we limit our approach to known constructs of OCL. The use of a limited subset of OCL to avoid undecidability is not new. For example $OCL-Lite$ [37] uses a limited subset of OCL to express constraints on UML class diagrams. Similarly, in our approach, we use a limited subset of OCL to express state invariants in statechart diagrams.

Hnatkowska and Huzar [38] analyzed the consistency of the statechart diagram of a class by writing OCL rules manually, and then execute the OCL rules by using an OCL compiler. This approach is limited to analyzing consistency of statechart diagrams against the class description and not using state invariants. In OCL, the model validation rules must be defined explicitly, based on the syntax of the UML models. In our approach, model validation is defined on the semantic interpretation of the OCL and UML models. The difference is that while OCL must define a large number of well-formed rules for different variations and combinations of model elements, a logic approach requires a smaller number of axioms that are often simpler.

To the best of our knowledge, none of the above-mentioned works proposed an automatic translation and consistency checking approach for UML models with state invariants.

## 3.2. Consistency of class and object diagrams

The use of ontology languages and description logic in the context of model validation has been proposed in the past by different authors [15–17,39–41,32]. However, to the best of our knowledge, none of them has addressed the reasoning of composition, ordered properties and non-unique properties in detail, neither the enforcement of the closed-world restrictions in OWL 2 DL. These works focus on the problem of ontology modeling or on class diagram satisfiability, i.e. if a class diagram can be instantiated or not.

The validation of UML models using OCL has been discussed by several authors, including [42,43]. In OCL, the model conformance rules must be defined explicitly based on the syntax of the UML models. In our approach, model conformance is defined on the semantic interpretation of the models. The difference is that while OCL must define a large number of well-formed rules for different variations and combinations of model elements, a logic approach requires a smaller number of axioms that are often simpler.

MOVA tool [44] provides the facility to draw and validate models against a subset of the UML metamodels. There are a number of limitations in this tool, firstly, this tool produces MOVA specific XMI of UML models, and due to this the models generated by using this tool are not readable by any other modeling tools and vice versa. Secondly, this tool only supports a limited subset of a UML metamodel and does not support the object and class diagram concepts such as ordered properties and ordered links, composition and non-unique associations and links. Lastly, this tool admittedly does not support the full OCL syntax.

Alloy [23] is a tool for the validation of a model against the metamodel. In order to use this tool we need to give a model and its metamodel as an input in the form of an Alloy script. There are some plugins available for example UMLtoAlloy [45], which can transform a UML model into the Alloy script, but the details about whether they can translate UML concepts such as composition, ordered properties and non-unique associations are missing.

UML Analyzer [46] used a text based rulebase for the analysis of UML models. Any missing translation rule of UML constraints during translation may lead the whole validation result to become false positive.

The TWOUSE approach [47] is focused on two areas: the first is the ontology development modeling [48], and the second is the validation of DSLs by using OWL 2 reasoners [34]. However, their work leaves out some of the important UML concepts such as composition including unshared and acyclicity constraints, open-world assumptions in the translation of class specializations and class memberships, non-unique associations, ordered properties, and the validation of basic textual constraints like OCL. Their work is mainly conducted in parallel with our work on the metamodel validation [49] i.e., during the year 2010. Moreover, their work on the validation is limited to the validation of DSLs and does not offer the validation of object diagrams against the class diagrams, nor analyzes the consistency of statechart diagrams with or without invariants.

### 3.3. Consistency of multiple UML diagrams

The problem of model merging has been discussed by several authors in the past. For instance, Lutz et al. [50] discussed the merging of models by humans. In their approach, the common model elements in models are calculated manually, and then models are merged using these common model elements. Moreover, the approaches presented in [51–53] discuss the difference and union of UML models. These approaches propose to take the union of all models in order to perform model merge. We have also used a similar approach while merging UML models by translating the union of all model elements of all models into a single ontology. To the best of our knowledge, none of the above-mentioned works does this using OWL 2. Moreover, they do not discuss the automatic discovery of inconsistencies, which occur due to the merging of different versions of a UML model.

## 4. UML class diagrams in OWL 2 DL

In this section, we firstly explain our understanding of UML class diagram concepts using logic and show how to translate these concepts into OWL 2. Secondly, we discuss different types of textual constraint languages that we treat in our approach. We also show the equivalent OWL 2 translations of a subset of these languages. This section is based on the work presented in [54].

The UML class diagrams are defined as a set of classes and their relationships in the form of generalizations and associations [2, p. 144]. The diagrams are used to express the static content or a structure of the system under development. Unfortunately, the semantics of UML are mostly specified semi-formally by means of a textual description [2]. The problem of the automatic validation of these diagrams necessitates the need of a formalization that can be understood by a reasoner. Therefore, in this section, we show the formalization and the corresponding OWL 2 translations of UML class diagram concepts that we address in our approach.

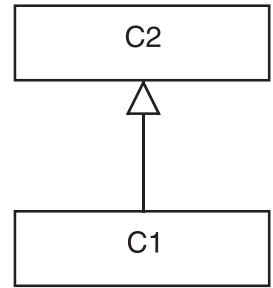

**Fig. 4.** Class specialization.

### 4.1. Class

A class in a class diagram represents a collection of objects which share the same features, constraints and definition. Each class in a class diagram is treated as a *class* in OWL 2. A UML class $C$ is translated in OWL 2 as

```
Declaration(Class(C))
```

### 4.2. Class specialization

Class specialization is reduced to the set inclusion. We represent the fact that a UML class C1 is a specialization of UML class C2 (see Fig. 4) with the condition:

$C1 \subseteq C2$

In this case, we say that C2 is a superclass of C1. If C2 is the superclass of C1 we say that they are in a specialization relation. The specialization relation $C1 \subseteq C2$ is translated in OWL 2 as

```
SubClassOf(C1 C2 )
```

### 4.3. Disjoint classes

We assume that an object cannot belong to two classes, except when these two classes are in a specialization relation. In our semantic interpretation of a UML class diagram, it is equally important to denote the facts that two classes are not in a specialization relation. This is due to the fact that in object-oriented models an object cannot belong to two classes, except when these two classes are in a specialization relation. We represent the fact that UML class C1 and UML class C2 are not in a specialization relation (see Fig. 5) with the condition:

$C1 \cap C2 = \varnothing$

With this condition, an object cannot belong to these two classes simultaneously. Due to the open-world assumption used in description logic, we need to explicitly state this fact in OWL 2 as

```
DisjointClasses(C1 C2 )
```

It is necessary for all classes to either explicitly or implicitly declare that they are disjoint from classes that they do not share model elements with. However, given that these axioms need to take the entire class hierarchy into account, how to efficiently generate this information is not immediately apparent. We have declared every class equivalent to the union of its subclasses and direct instances – or if it has no subclasses, equivalent to its direct instances. We have therefore provided enough information for a reasoner to be able to deduce which direct instances any given class is made up of. Correctly declaring the direct instances disjoint is consequently enough information for a reasoner to infer if any given pair of classes are disjoint.

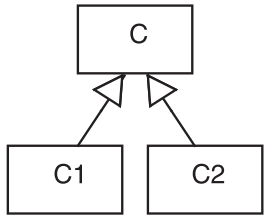

**Fig. 5.** Disjoint classes.

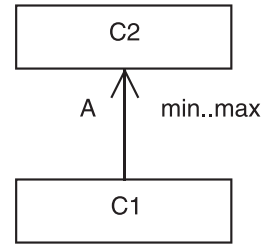

**Fig. 6.** Association.

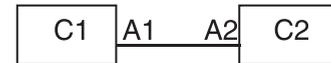

**Fig. 7.** Association bidirectionality.

A pair of classes are disjoint if neither is a superclass to or a subclass of the other, and they do not share any subclasses. The direct instance class never has any subclasses, so deciding whether it is disjoint to another class merely requires that we verify that the class in question is not a superclass of the direct instance. This make it necessary to only traverse part of the class hierarchy. Furthermore, this approach limits the amount of generated axioms to one per class. As classes inherit the properties of superclasses, it is only necessary to include top level classes and its superclasses' unshared direct subclasses in the axiom.

### 4.4. Associations

The association is another fundamental concept of UML class diagrams and it represents a basic relationship between the instances of two or more classes. We represent a UML directed binary association $A$ from class C1 to C2 (see Fig. 6) as a relation:

$A: C1 x C2$

Each association in a class diagram contains two properties, namely *Domain* and *Range*, that represent each end of the association. In our example, $C1$ is the domain and $C2$ is the range of the association $A$.

A UML association $A$ from UML class $C1$ to $C2$ is represented in OWL 2 as

```
Declaration(ObjectProperty(A))
ObjectPropertyDomain(A C1 )
ObjectPropertyRange(A C2 )
```

### 4.5. Multiplicity

A UML association in a class diagram is annotated with a positive number; this number indicates the multiplicity of an association. Association multiplicity describes the number of allowable objects of a range class to link with the object of a domain class. The multiplicity of an association defines additional conditions over this relation:

$\#\{y|(x,y) \in A\} \geq min$

$\#\{y|(x,y) \in A\} \leq max$

We map the multiplicity of a UML association into OWL 2, by defining the domain class of an association as a subclass of a set of classes, which relates with the same property and the given cardinality.

The UML association $A$ from class $C1$ and $C2$ having a multiplicity constraint of ($min$,$max$) is represented in OWL 2 as

```
SubClassOf(C1 ObjectMinCardinality(min A ) )
SubClassOf(C1 ObjectMaxCardinality(max A ) )
```

### 4.6. Bidirectionality

In UML, the associations that share opposite domain and range form a bidirectional association. For example if $A1$ and $A2$ in Fig. 7 are UML associations and both associations share opposite domain and range, then both associations are considered as bidirectional of each other, such that

$A1 = \{(x,y)|(y,x) \in A2\}$

The UML bidirectionality between the associations $A1$ and $A2$ is expressed in OWL 2 as

```
InverseObjectProperty(A1 A2 )
```

### 4.7. Association generalization

An association can be generalized by another association. The association generalization is also known as association subsetting. Association subsetting allows the specialization of an existing association, with new characteristics while retaining its existing features, such as domain and range. However, we can reassign a domain and a range of a subassociation, provided that the new domain and the range of a subassociation are the subclasses of the domain and the range of a parent association. Each instance of the specialized association is also an instance of the original property. Therefore, elements that are a part of its slot should be a part of the original association slot. The association subsetting between association $A1$ and $A2$ shown in Fig. 8 is defined as

$$A1 \subseteq A2$$

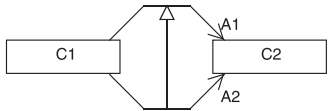

**Fig. 8.** Association generalization $A1 \subseteq A2$.

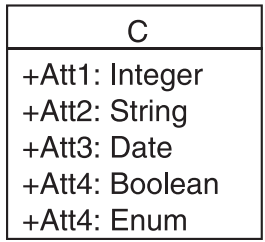

**Fig. 9.** Class attributes.

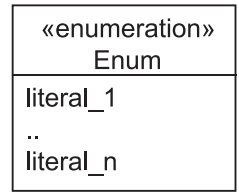

**Fig. 10.** Enumeration datatype.

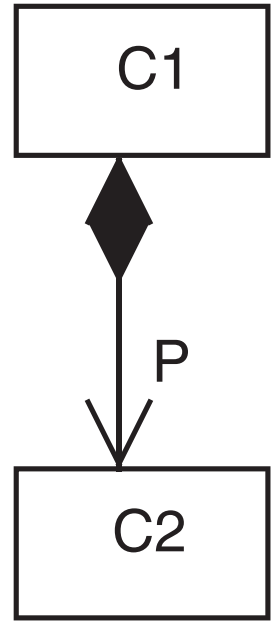

**Fig. 11.** Composition C1 owns C2.

Where $A1$ is a subassociation of $A2$, and translated in OWL 2 as

```
SubObjectPropertyOf(A1 A2 )
```

### 4.8. Class attributes

The class attributes (see Fig. 9) depicting variables of datatypes such as string, integer or boolean are also considered as relations. In this case, the range of the relation $A$ belongs to the set $D$ representing the datatype as

$$\forall x, y : (x, y) \in A \Rightarrow y \in D$$

Attributes usually have a multiplicity restriction to one value. The attributes of a UML class in a class diagram are translated in OWL 2 as a *DataProperty*. In OWL 2, the data property uses datatype in its range. The datatype can be xsd: boolean, xsd:string, xsd:int and other datatypes (shown in [19, Table 3]). We map attributes that use basic types by declaring a data property with the attribute's name. An attribute is a required component of its class. Consequently, the data properties describing attributes have an exact cardinality of one. The attribute *Att* of the UML class *C* having any of the above-mentioned *DataType* is translated in OWL 2 as

```
Declaration(DataProperty(Att ))
SubClassOf(C DataExactCardinality(1 Att ))
DataPropertyDomain(Att C ) DataPropertyRange(Att
    DataType )
```

### 4.9. Data enumeration

Enumeration is a kind of datatype, whose instances are user-defined enumeration literals [2, p. 67]. The enumeration

*Enum* (Fig. 10) is declared by using a `DatatypeDefinition` axiom in OWL 2 DL. The class attribute *Att* having a datatype *Enum*, means

$$\forall x, y : (x, y) \in Att \Rightarrow y \in Enum$$

where *Enum* being a set of enumeration literals $\{(literal_1), \ldots, (literal_n)\}$ is represented in OWL 2 as

```
DataPropertyRange(Att
DataOneOf()literal1) ^ ^ datatype..))
```

### 4.10. Composition

In composition, an object of a class is made up of parts that are the objects of another class. To give a formal definition of composition, we use a single predicate *owns* to keep track of the composition relationships. If C1 owns C2 via a composition association $P$ as shown in Fig. 11, and $P$ is the property from C1 to C2, then

$$\forall x, y : (x, y) \in P \Rightarrow x \in C1$$
$$\forall x, y : (x, y) \in P \Rightarrow y \in C2$$

$$P \subseteq owns$$

Composition relationships are defined in UML by two constraints, exclusive ownership and acyclicity. Exclusive ownership means that an object can have only one owner:

$$\forall x, y, z : (x, z) \in owns \quad \text{and} \quad (y, z) \in owns \Rightarrow x = y$$

Acyclicity means that an object cannot transitively become an owner of itself. A situation where an object $x$ owns $y$, $y$ owns $z$ and $z$ owns $x$ is disallowed. A necessary and sufficient condition for acyclicity of *owns* is that the transitive closure of the relation is irreflexive. We can define the transitive closure of *owns*, in the following way:

$$\forall x, y, z : (x, y) \in owns \quad \text{and} \quad (y, z) \in owns \Rightarrow (x, z) \in owns$$

Irreflexivity of the transitive closure is then simply expressed as

$$\forall x : x \in \Delta_i \Rightarrow (x, x) \notin owns$$

In order to translate the composition association mentioned above into OWL 2, we first define an object property named "P" from class $C1$ to class $C2$.

```
Declaration(ObjectProperty(P))
ObjectPropertyDomain(P C1)
ObjectPropertyRange(P C2)
```

Next, we consider the exclusive ownership constraint of the composition on the owning end of a composite relationship. To implement the single owner requirement of a composition relationship in OWL 2, we have firstly defined the global property *owns* as

```
InverseFunctionalObjectProperty(owns)
```

The inverse functional property will restrict the individuals of containing class to link with more than one individuals of owning class. Secondly, we make the composite relationship "P", a subproperty of the global property *owns*

```
SubObjectPropertyOf(P owns)
```

Moreover, in order to capture the acyclic requirement of the composition, we make *owns* transitive and irreflexive at the same time. Transitivity will capture the self ownership and irreflexiveness will disallow the individual becoming an owner of itself. This is equivalent to saying that the transitive closure of the ownership property is irreflexive. However, it is not possible to combine a cardinality restriction with transitive properties [55]. In OWL 2 DL, if we could do so, the logic system would no longer be decidable, and we would not be able to use a fully automatic reasoner to carry out validation. To solve this problem, we translate transitivity in Semantic Web Rule Language (SWRL) and irreflexivity in OWL 2. The transitivity of the property $\Rightarrow$ *owns* is written in SWRL as

$$owns(?x, ?y) \land owns(?y, ?z) \Rightarrow owns(?x, ?z)$$

and irreflexivity of *owns* is translated in OWL 2 as

```
IrreflexiveObjectProperty(owns)
```

## 5. Class diagrams including OCL constraints

In this section, we show the formalization and translation of OCL constraints that can be applied on the UML

diagrams, and also motivate our choice of features that are included in the formalizations.

The UML specification proposes the use of Object Constraint Language (OCL) [56] to define constraints on UML models, such as the restrictions on the values of object attributes and the restrictions on the existence of objects by using a multiplicity constraint of an association. These constraints are combined using boolean operators.

The OCL constraints may also have inconsistencies. According to Wilke and Demuth [57], 48.5% of the OCL constraints used for expressing the well-formedness of UML in OMG documents are erroneous. The erroneous OCL constraints may cause a context class in a class diagram to become unsatisfiable. In order to identify the inconsistencies in OCL constraints, we need to do the reasoning of these constraints. But unfortunately, in general, OCL is not decidable. However, we can avoid undecidability by restricting our approach to a reduced fragment of the full OCL. Therefore, in order to use the reasoners for checking the inconsistencies in OCL constraints, we use a reduced subset of OCL which is limited to the constructs of multiplicity, attributes value and boolean operators. In this section, we discuss and translate the different types of OCL constructs supported in our approach. The grammar of OCL supported in our approach is shown in Fig. 12.

### 5.1. Linking OCL constraints with classes in OWL 2

Each OCL constraint of a class diagram comprises a number of elements such as context, name and the invariant. These elements hold the name of a constrained class, the name of an invariant and the constraint in OCL, respectively. The OCL constraints are used to apply restrictions on object memberships, i.e. an object can only belong to a particular class if it fulfills the conditions applied by the OCL constraint of that class, so that

$$Context \equiv Invariant$$

### 5.2. Attribute constraints

The value of the attribute is accessed in OCL by using a keyword *self* or by using a class reference [56, p. 15], the value constraint of the attribute *Att* is written in OCL as `self.Att=Value`, meaning

$$\{x | (x, Value) \in Att\}$$

where *Value* represents the attribute value. The restriction on the value of the attribute is translated in OWL 2 by using the axiom `DataHasValue`. The OCL attribute value constraint `self.Att=Value` is translated to OWL 2 as

```
DataHasValue(Att }Value}^^datatype)
```

In this translation, *Att* is the name of the attribute, *Value* is the value of the attribute, and *datatype* is the datatype of the attribute *Value*.

$$
\begin{aligned}
\langle\text{OCL-expression}\rangle &::= \quad \langle\text{cond-expr}\rangle \ (\langle\text{logic-op}\rangle\langle\text{cond-expr}\rangle)^* \\
\langle\text{logic-op}\rangle &::= \quad and \mid or \\
\langle\text{cond-expr}\rangle &::= \quad \langle\text{ref}\rangle \rightarrow\text{size}()\langle\text{relational-operator}\rangle\langle\text{integer-literal}\rangle \\
&\quad \mid \ \langle\text{ref}\rangle\langle\text{relational-operator}\rangle\langle\text{primitive-literal}\rangle \\
&\quad \mid \ \langle\text{ref}\rangle \rightarrow\text{isEmpty}() \mid \langle\text{ref}\rangle \rightarrow \text{notEmpty}() \\
&\quad \mid \ \langle\text{ref}\rangle \rightarrow\text{excludes}(\langle\text{ref}\rangle) \\
\langle\text{ref}\rangle &::= \quad \text{self.}\langle\text{identifier}\rangle \\
\langle\text{identifier}\rangle &::= \quad '\{\langle\text{characters}\rangle\} \mid 0..9 \ \{0..9\}' \\
\langle\text{relational-operator}\rangle &::= \quad < \mid <= \mid > \mid >= \mid <> \mid = \\
\langle\text{primitive-literal}\rangle &::= \quad \langle\text{boolean-literal}\rangle \mid \langle\text{integer-literal}\rangle \\
&\quad \mid \ \langle\text{string-literal}\rangle \mid \text{null} \\
\langle\text{boolean-literal}\rangle &::= \quad \text{true} \mid \text{false} \\
\langle\text{integer-literal}\rangle &::= \quad 0..9 \ \{0..9\} \\
\langle\text{string-literal}\rangle &::= \quad '\{\langle\text{characters}\rangle\}'
\end{aligned}
$$

**Fig. 12.** The grammar of the supported OCL fragment.

### 5.3. Multiplicity constraints

The multiplicity of an association is accessed by using the *size*() operation in OCL [56, p. 144]. The multiplicity constraint on the association *A* in OCL is written as `self.A−>size()=Value`, where *Value* is a positive integer and represents the number of allowable instances of the range class of the association *A*. We can use a number of value restriction infix operators with the *size*() operation, such as $=, \ >=, \ <=, \ <$ and $>$. The multiplicity constraint on an association *A* is defined as

$$\{x \mid \#\{y \mid (x,y) \in A\} \, OP \, Value\}$$

where *OP* is the infix operator and *Value* is a positive integer. The translation of the *size*() operation in OWL 2 is based on the infix operator used with the *size*() operation, such as "*size*() $>=$" or "*size*() $>$" translated using the OWL 2 axiom ObjectMinCardinality, "*size*() $<=$" or "*size*() $<$" translated using the OWL 2 axiom ObjectMaxCardinality, and "*size*() $=$" translated using the OWL 2 axiom Object Exact Cardinality.

For example, `self.A−>size()=Value` is an OCL constraint, in which *A* is the name of an association and *Value* is a positive integer, is translated to OWL 2 as

    ObjectExactCardinality(Value A)

Furthermore, the constructs *isEmpty* and *notEmpty* represent *size*() $=0$ and *size*() $>0$ respectively. The invariant `self.A−>isEmpty()` is translated in OWL 2 as

    ObjectExactCardinality(0 A)

and similarly the invariant `self.A−>notEmpty()` is translated in OWL 2 as

    ObjectMinCardinality(1 A)

The OCL constraints over the multiplicity of an association are further extended by using construct *excludes*. This construct is used to apply restriction on the objects of a domain class of an association to not to link with some specific objects of a range class. For example, if we have an association *A* with domain and range class *C* and there is a condition that an object of class *C* cannot link with itself by using a link of an association *A*, such that `self.A−> excludes(self.A)`, the construct *excludes* in this specific case is translated in OWL 2 as

    IrreflexiveObjectProperty(A )

In OWL 2, an association that is declared *Irreflexive* disallows an object of its domain or range class to link with itself.

### 5.4. Boolean operators

The constraints in a state invariant are written in the form of a boolean expression, and joined by using the boolean operators, such as "*and*" and "*or*" [56, p. 144].

The binary "*and*" operator evaluates to true when both boolean expressions $Ex_1$ and $Ex_2$ are true. In our translation, this is represented by the intersection of the sets that represent both expressions, i.e.

$$Ex_1 \cap Ex_2$$

This is represented in OWL 2 as

    ObjectIntersectionOf(Ex1 Ex2)

The binary "*or*" operator evaluates to true when at least one of the boolean expression $Ex_1$ or $Ex_2$ is true. In our translation, this is represented by the union of the sets that represent both expressions, such as

$Ex_1 \cup Ex_2$

This is represented in OWL 2 as

```
ObjectUnionOf(Ex1 Ex2)
```

### 5.5. Example

The OCL invariant $self.hasParent - > excludes(self.hasParent)$ has the context class *Person* shown in Fig. 13 representing a condition that a person cannot become a parent of itself. The class diagram with the OCL constraint is translated in OWL 2 as

```
Declaration(Class(Person))
Declaration(ObjectProperty(hasParent ))
ObjectPropertyDomain(hasParent Person)
ObjectPropertyRange(hasParent Person)

IrreflexiveObjectProperty(hasParent)
```

The details about how the inconsistent OCL constraint makes a UML model inconsistent and how to detect the inconsistencies in OCL constraints using OWL 2 reasoners and the conformance of the objects against the OCL constraints specified with a class diagram are discussed later in this paper.

## 6. UML object diagrams

In this section we give a formal definition of the UML class and object diagraming concepts that we treat in our approach. The definition is given in terms of predicate logic. In this section we also give the translation of UML Object diagraming concepts to OWL 2 and motivate our choice of features that are included in the definition.

### 6.1. UML classes and objects

A UML class represents a set of objects that have the same characteristics [2]. A UML class C is defined as a unary predicate $C$ in predicate logic.

An instance of a class is called an object. In UML, every object in an object diagram must belong to a specific class in a class diagram. An object $x$ in an object diagram that belongs to a class C in a class diagram is defined in predicate logic as

$C(x)$

A UML Class C in the class diagram is translated to OWL 2 as

```
Declaration(Class(C ) )
```

Furthermore, every object that exists in an object diagram must belong to a specific class in a class diagram.

In OWL 2 the UML object is represented as a class assertion and is called an *individual*. A UML object x of the class C is translated in OWL 2 as

```
ClassAssertion(C x )
```

Furthermore, every object in a UML object diagram is by default different from another. Whereas, in OWL 2 due to the open-world assumption, we need to explicitly mention that all individuals are different from each other. For example: for objects $x1,..,xn$ in an object diagram, we use the OWL 2 axiom:

```
DifferentIndividuals(x1..xn)
```

### 6.2. Class memberships

The UML and MOF semantics for instantiation follow closely the object-oriented paradigm i.e. the closed world assumptions [58]. In this context, when declaring that a model element m is of type C, it is required that the following holds:

1. m is a direct instance of the class C.
2. m is an indirect instance of all the superclasses of C.
3. m is not an instance of any other class.

However, OWL 2 follows the open world assumption. Therefore, declaring that an element m is of type C in OWL 2 asserts that

1. m is a direct or indirect instance of the class C.
2. m is an indirect instance of all the superclasses of C.
3. m is not an instance of any classes explicitly declared disjoint to C.

Furthermore, these semantic differences require that we introduce additional axioms in our ontology to restrict object membership to the intended class hierarchies.

### 6.3. Class memberships within inheritance hierarchies

According to UML semantics of class membership, whenever we assert that a model element is an instance of a class, we also assert that it is not an instance of its subclasses. But in OWL 2 this is not the case. In order to overcome this semantic gap between OWL 2 and UML, we translate UML class into OWL 2 as a union of two disjoint concepts.

```
EquivalentClasses(C
  ObjectUnionOf(C_Direct C1..Cn ) )
DisjointClasses(C_Direct
  ObjectUnionOf(C1..Cn ) )
```

First concept *C_Direct*, represents the collection of all the objects which are direct instances of a class. And, the second concept represents all objects that belong to its subclasses $C1,..,Cn$.

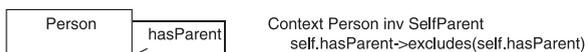
Context Person inv SelfParent
    self.hasParent->excludes(self.hasParent)

**Fig. 13.** A UML class diagram with OCL constraints.

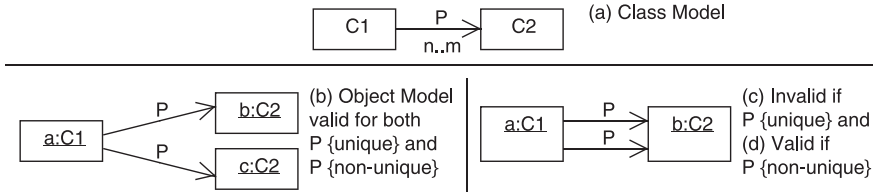

**Fig. 14.** (a) A class diagram depicting an association P connecting two classes, (b) a consistent object diagram for both unique and non-unique P, (c) an inconsistent object diagram if P is unique, and (d) a consistent object diagram if P is non-unique.

### 6.4. UML association and links

A UML binary association defines a relationship between two classes [2]. A UML link is an instance of an association. A link l of an association P connecting objects x and y is represented in predicate logic as

$P(x, y, l)$

We often do not need to differentiate what link is used to connect two objects. Therefore it is convenient for us to define the following:

$\forall x, y, l. P(x, y, l) \rightarrow P(x, y)$

A UML association is represented in OWL 2 as an object property. An association P from a class C1 to a class C2 is represented in OWL 2 as

```
Declaration(ObjectProperty(P))
ObjectPropertyDomain(P C1)
ObjectPropertyRange(P C2)
```

A link in OWL 2 is represented as a property assertion. The link of an association P between the objects x1 and x2 in an object diagram is represented in OWL 2 as

```
ObjectPropertyAssertion(P x1 x2)
```

Moreover, due to the open-world assumptions of OWL 2, for a reasoner to be able to detect a violation of a minimum multiplicity constraint, we need to provide a definitive knowledge about the links, connecting or not connecting the individuals of a domain class and a range class of an association. Therefore, if there is no link between the objects of a domain class and a range class of a UML association, we need to explicitly declare that there is no connection between the individuals. The knowledge about the non-existence of a link between individuals is called a negative assertion. The negative assertion of an association P between the objects x1 and x2 in OWL 2 is written as

```
NegativeObjectPropertyAssertion(P x1 x2)
```

A negative assertion is required when there exists an association between the classes but their specific individuals are not connected with a link.

### 6.5. Unique and non-unique associations

A UML association multiplicity can be unique or non-unique. A unique association does not allow two objects to be linked with each other more than one time, as shown in Fig. 14c.

The restriction of unique association P is written in predicate logic as

$\forall x, y, l_1, l_2. P(x, y, l_1) \land P(x, y, l_2) \rightarrow l_1 = l_2$

In the case of non-unique associations this restriction does not apply.

In OWL 2 the UML unique association is treated as a normal object property, and the translation of a multiplicity constraint of a unique association is done by using OWL 2 axioms: ObjectMinCardinality and Object Max Cardinality.

However, in case of a non-unique association, there can be multiple links between the objects of a domain class and a range class, as shown in Fig. 14d. The OWL 2 reasoner considers all links which have a common source and target as one link, and to make the reasoner able to consider all those links as different links, we have introduced an intermediate class in between a domain class and a range class of a non-unique association. As a consequence, every non-unique association P is translated in OWL 2 as a combination of two object properties $P\_I$ and $I\_P$. Where $P\_I$ connects a domain class to an intermediate class, and $I\_P$ connects an intermediate class to a range class of a non-unique association. The object property $P\_I$ is written in OWL 2 as

```
InverseFunctionalObjectProperty(P_I)
```

An inverse functional object property restricts an individual of a domain class connecting with more than one individuals of an intermediate class. Furthermore, we put a cardinality restriction of n..m on $P\_I$ by using OWL 2 axioms: ObjectMinCardinality and ObjectMaxCardinality. Moreover, the object property $I\_P$ is written in OWL 2 as a normal object property. In order to ensure that the individuals of an intermediate class $C\_I$ connect one to one with the individuals of a range class, we have to put the exact cardinality of one on the property $I\_P$ as

```
SubClassOf(C_I ObjectExactCardinality(1 I_P))
```

Furthermore, a link of a non-unique association P from object a to b in an object diagram is translated as the assertions of the object properties $P\_I$ and $I\_P$ as

```
ObjectPropertyAssertion(P_I a C_I_#)
ObjectPropertyAssertion(I_P C_I_# b)
```

where $C\_I\_\#$ is an individual of an intermediate class $C\_I$, and # is an auto generated unique number which is responsible to create a distinction between the identical links of a non-unique association. For example, the translation of identical links of the object diagram shown in Fig. 14d is

```
\\ Link 1
ObjectPropertyAssertion(P_I a C_I_1 )
ObjectPropertyAssertion(I_P C_I_1 b )
\\ Link 2
ObjectPropertyAssertion(P_I a C_I_2 )
ObjectPropertyAssertion(I_P C_I_2 b )
```

### 6.6. Ordered properties

In UML ordering, the links of an ordered property are labeled with a unique numbered index, and it is required that the indexes are in order. An ordering can be used for example to preserve a sequence of a parameter in a function. A link of an ordered property $P$ connecting object $x$ of the source class and object $y$ of the target class of the ordered property $P$ having a label $i$ is represented in predicate logic as

$$P(x, y, i)$$

Furthermore, all links of an ordered property having identical index $i$ are required to have an identical source and target, in predicate logic it is represented as

$$\forall i \in N \forall x, y, a, b.P(x, a, i) \land P(y, b, i) \rightarrow (x = y) \land (a = b)$$

where, $x, y$ are the objects of a domain class and $a, b$ are the objects of a range class of an ordered property $P$.

The UML ordered property is translated in OWL 2 as a normal ObjectProperty. The translation of basic constraints like domain, range and multiplicity is also the same as mentioned in Section 6.4. For example the translation of an ordered property $P$ depicted on top of Fig. 15 is as follows.

```
Declaration(ObjectProperty(P ) )
ObjectPropertyDomain(P C1 )
ObjectPropertyRange(P C2 )
SubClassOf(C1 ObjectMinCardinality(n P ) )
SubClassOf(C1 ObjectMaxCardinality(m P ) )
```

Moreover, in a UML object diagram, a link of an ordered property is labeled with an index, and requires that the index is unique and in order. In OWL 2 there is no specific axiom for the representation of a UML ordered property link or any link with a label. Due to this fact, we translate a UML ordered property link into OWL 2 in four steps. First, declare an *index* property for each ordered property link that exist in an object diagram. The *index* property is declared in OWL 2 as

```
Declaration(ObjectProperty(index_P_# ) )
```

To make every index property link reachable while parsing a translated object diagram in OWL 2, the name of an *index* property comprises three parts. First, *index* refers that the link of this object property will represent an ordered property link. Second, $P$ refers the name of an ordered property in a class diagram. Third, # is representing an index or a label on a link of an ordered property. The data type of an index can be "xsd:integer" [59] or "xsd:string" [59]. For example, for each labeled link 1 and 2 of the ordered property $P$ shown at the bottom of Fig. 15, we will declare an *index* property in OWL 2 as

```
Declaration(ObjectProperty(index_P_1 ))
Declaration(ObjectProperty(index_P_2 ))
```

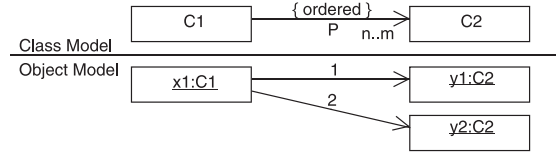

**Fig. 15.** Top: a UML class diagram depicting ordered property. Bottom: a UML object diagram depicting ordered links.

Second, a domain and a range of each *index* property in OWL 2 is the same as the domain and the range of an ordered property. Therefore, each index property in OWL 2 will be a subproperty of an ordered property $P$:

```
SubObjectPropertyOf(index_P_# P )
```

Third, all links of an *index* property must have an identical source and target. In OWL 2 all links having an identical source and target are considered as one link. Therefore, we have also made the *index* property:

1. FunctionalObjectProperty so that one link of an *index* property may not lead to two individuals.
   `FunctionalObjectProperty(index_P_# )`
2. InverseFunctionalObjectProperty so that two links of an *index* property may not lead to one individual.
   `InverseFunctionalObjectProperty(index_P_# )`

Last, each *index* property in OWL 2 is instantiated among the respective individuals of a domain and a range class of an ordered property. For example the UML ordered property links $P(x1, y1, 1)$ and $P(x1, y2, 2)$ as shown at the bottom of Fig. 15, is represented in OWL 2 as

```
ObjectPropertyAssertion(index_P_1 x1 y1)
ObjectPropertyAssertion(index_P_2 x1 y2)
```

## 7. UML model merge using OWL 2

A software development project involves the creation of many models. These models may represent different versions of the same software component, often designed in parallel by a number of designers. These different versions of a model may create contradictions when combined. This raises a need of a mechanism to semantically merge different versions of a UML model together and find out the possible contradictions and inconsistencies arise between model elements when they are viewed together.

The issue of merging UML models has been studied in the research literature previously [50–53]. However, these approaches do not provide the mechanism to check the consistency of merged models. In this section, we propose an approach to study how different models of the same metamodel are merged, and later in this paper we show how to validate the merged models. In the validation of merged models, we want to identify the possible inconsistencies that arise when the different models of the same metamodel are viewed together/merged. In this approach we use a decidable fragment of Web Ontology Language (OWL 2 DL) [19] to represent and merge UML models, and then use OWL 2 reasoning tools to determine the inconsistencies in the merged model.

### 7.1. Model merging using OWL 2

In this section we show how to perform model merging. The merging of given models is performed by putting the union of all model elements of all models in to a single model, i.e. the merged model, so that, if $M1$ and $M2$ are given models then the merged model $M$ represents the union of all model elements of all given models, i.e., $M = M1 \cup M2$. For example, in Fig. 16 the merged model $M$ is a union $(M1 \cup M2)$ of given models $M1 = \{A, C, B\}$ and $M2 = \{A, C, D\}$.

In order to merge the UML models representing different versions of a UML model, we propose to use description logic [18]. Furthermore to detect the inconsistencies originating from the merging of different versions of a model, we propose to use the automated reasoning tools [20,21]. A number of UML models representing different versions of a UML model are taken as an input. All the inputs are translated to OWL 2 DL and then merged into a single ontology. Then the ontology is passed to a reasoner that produces a validation report. The validation report reveals the inconsistencies in the ontology representing a merged UML model.

In order to demonstrate the merging of different versions of a UML model, we first translate all UML models into OWL 2 DL by using the method discussed in previous sections and merge the OWL 2 translations of all UML models into a single ontology. Since we translate all model elements of all UML models into a single ontology, the common elements of all models (i.e. $M1 \cap M2$ in our example Fig. 16) will overlap. Additionally, due to the open-world assumptions [58] of OWL 2, where model elements are recognized by their names, all model elements having the same name are considered as a single model element or concept. Consequently, due to this assumption all models will merge or connect with each other by using common model elements. In Fig. 17 we provide an example of the OWL 2 translation of models $M1$ and $M2$ given earlier in Fig. 16.

The given models $M1$ and $M2$ represent six classes in total i.e., $M1 = \{A, C, B\}$ and $M2 = \{A, C, D\}$. We translate all six classes of $M1$ and $M2$ into a single ontology. Due to the unique name assumption of OWL 2, the reasoner recognises distinct classes, and counts all six classes, i.e, $M1 = \{A, C, B\}$ and $M2 = \{A, C, D\}$ as four classes, i.e., $M1 \cup M2 = \{A, C, B, D\}$.

The unique name assumption of OWL 2 is also applied on the relationships such as associations, generalization and on association constraints such as multiplicity, composition, domain and range constraints. For example, both models $M1$ and $M2$ represent four relationships in total, in which there are two associations and two generalization relationships. Both models depict association $P$ from the class $A$ to the class $C$, therefore, due to the unique name assumption of OWL 2, the reasoner recognises the distinct associations and count three relations instead of four. Within these relations there are two generalization relationships and one association.

### 7.2. Classification report

In order to determine the number of elements in an ontology, the reasoner produces a classification report of

```
//M1

Declaration(Class(A))

Declaration(Class(C))

Declaration(Class(B))

SubClassOf( B A )

Declaration(ObjectProperty( P ))

ObjectPropertyDomain(P A)

ObjectPropertyRange(P C)

//M2

Declaration(Class(A))

Declaration(Class(C))

Declaration(Class(D))

SubClassOf( D C )

Declaration(ObjectProperty( P ))

ObjectPropertyDomain(P A)

ObjectPropertyRange(P C)
```

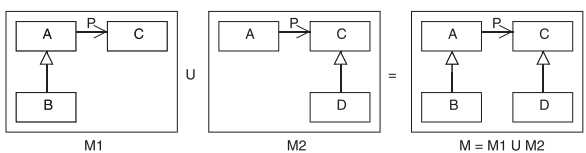

**Fig. 16.** The merge or union of two versions of a model.

**Fig. 17.** The OWL 2 ontology of models $M1$ and $M2$ shown in Fig. 16.

an ontology. The classification report of the ontology shown in Fig. 17 is given in Fig. 18. The classification report clearly shows that the reasoner recognises the distinct model elements of all models, i.e., $M1 \cup M2 = \{A, C, B, D\}$. It also shows the total number of seven model elements found in the ontology, which includes four classes $\{A, C, B, D\}$, two subclass relationships, and one association.

## 8. UML statechart diagrams to OWL 2 DL

A statechart diagram provides the behavioral interface of a class and defines the sequence of method invocations, the conditions under which they can be invoked and their expected results. In order to analyze the satisfiability of state invariants in a statechart diagram, we need to translate the states and their invariants into OWL 2 DL. The translation of the state and the state invariant includes the reference of the class and its attributes. Therefore,we translate a statechart diagram in the same ontology that contains the OWL 2 translation of a class diagram.

### 8.1. State and state hierarchy

We represent a UML state as a concept representing the objects that have such state active. A concept representing a state will be included in the concept representing all object instances of the class associated to the statechart diagram, since all objects that can have the state active belong to the given class. That is, if the state $S$ belongs to a statechart diagram describing the behavior of the class $C$, then

$$S \sqsubseteq C$$

We represent this in OWL 2 as follows:

```
Declaration(Class(S))
SubClassOf(S C )
```

State hierarchy is also represented using the concept inclusion. Whenever a substate is active, its containing state is also active. This implies that the concept representing a substate will be included in the concept representing its parent state, such as

$$sub \sqsubseteq S$$

This is represented graphically in Fig. 19 and translated in OWL 2 as

```
SubClassOf(sub S )
```

```
D:\pellet-2.3.0>pellet classify -l OWLAPI D:\Merging.owl2fs

Classifiying 7 elements

Classifiying:  100% complete in 00:00

Classifiying finished in 00:00

owl:Thing
   merging.owl2fs:A
      merging.owl2fs:B
   merging.owl2fs:C
   merging.owl2fs:D
```

**Fig. 18.** The classification report of the OWL 2 ontology shown in Fig. 17 generated by Pellet.

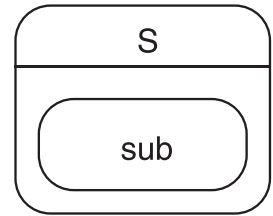

**Fig. 19.** State and state hierarchy.

### 8.2. Non-orthogonal states are exclusive

The UML superstructure specification requires that if a composite state is active and not orthogonal, at most one of its substates is active ([2], p. 564). This means that an object cannot be at the same time in the two concepts representing two exclusive states, i.e., if $S_1$ and $S_2$ represent substates of an active and not orthogonal composite state $S$ (see Fig. 20) then:

$$S_1 \sqcap S_2 = \bot$$

When representing a statechart diagram in OWL 2,the nonorthogonal exclusive states are declared as disjoint,so that they may not be able to share any object.

```
DisjointClasses(S1..Sn )
```

### 8.3. Orthogonal states are non-exclusive

The UML superstructure specification requires that if a composite state is active and orthogonal, all of its regions are active [2, p. 564]. That is if $R_1$ and $R_2$ are concepts representing the two regions of an orthogonal composite state represented by the concept $S$ (see Fig. 21) then

$$R_1 \sqcup R_2 = S$$

We should note that if $S_1$ and $S_2$ represent two substates where

$$S_1 \sqsubseteq R_1$$
$$S_2 \sqsubseteq R_2$$

then they are not exclusive and

$$S_1 \sqcap S_2 \neq \bot$$

Due to the open-world assumption of DL, concepts may represent common individuals unless they are explicitly declared as disjoint.

### 8.4. State invariant into OWL 2 DL

The UML specification defines a state in a UML diagram as the representation of a specific condition "A state models a situation during which some (usually implicit) invariant condition holds" [2, pp. 559–560]. We understand from this definition that the invariant condition characterizes the state: if the invariant condition holds the state is active, otherwise if the invariant condition does not hold the state is not active.

In our approach we represent an invariant (see Fig. 22) as an OWL 2 concept representing objects that make that invariant evaluate to true. Since the invariant holds iff the associated state is active, the concept representing a state will be the same as the concept representing an invariant. This is represented in OWL 2 as an equivalent class relation between the state $S$ and its invariant:

```
EquivalentClasses (S Invariant)
```

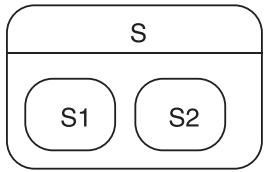

**Fig. 20.** Non-orthogonal states are exclusive.

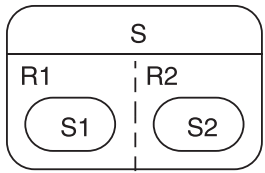

**Fig. 21.** Orthogonal states are non-exclusive.

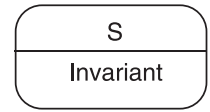

**Fig. 22.** OCL state invariant.

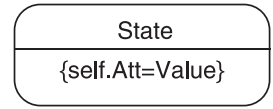

**Fig. 23.** OCL state invariant – attribute constraint.

Due to the equivalent relationship between the state and its invariant, all objects that fulfill the condition of its state invariant will also be in that specific state.

*State constraints*: The UML also allows us to define additional constraints to a state, and names these constraints also as state invariants. However,the semantics of a state constraint are more relaxed since it "specifies conditions that are always true when this state is the current state" [2, p. 562]. In this sense, the state constraints define necessary conditions for a state to be active, but not sufficient. This means that the actual state invariant may remain implicit. However, we consider a state invariant as a predicate characterizing a state. That is, a state will be active if and only if its state invariant holds.

*A state invariant characterizes a state*: The UML super-structure specification requires that whenever a state is active its state invariant evaluates to true [2, p. 562]. A consequence of this is that state invariants should be satisfiable. That is, every state invariant in a statechart diagram must hold in at least one object configuration. Otherwise there cannot be objects that have such state active. Since invariants should be satisfiable, the concept $S$ representing a state should be satisfiable, i.e.

$S \neq \bot$

### 8.5. OCL state invariants to OWL 2 DL

A state invariant is a runtime constraint on a state in a statechart [2, p. 514]. The UML specification proposes the use of OCL to define constraints in UML models, including state invariants. OCL is well supported by many modeling tools [60,61]. Unfortunately, in general OCL is not decidable. However, we can avoid undecidability by restricting our approach to a reduced fragment of the full OCL [62]. The use of a limited fragment of OCL to avoid undecidability has been proposed in the past also by other authors [62,37].

In this paper, we consider OCL constructs using mainly multiplicity, attributes value and boolean operators. The detail translation of OCL to OWL 2 has been already discussed in Section 5.5. The only difference is the context of OCL constraint. In case of class diagrams the context of an OCL constraint is a *Class*, whereas in case of statecharts the context of OCL constraint is a *State*, for example:

*In case of Attribute Constraints*: The restriction on the value of the attribute is translated in OWL 2 by using the axiom `DataHasValue`. The OCL attribute value constraint `self.Att=Value` (see Fig. 23) is translated in OWL 2 as

```
EquivalentClasses (State
  DataHasValue(Att }Value}^^datatype))
```

In the above translation, *State* is the context of the OCL constraint, *Att* is the name of the attribute, *Value* is the value of the attribute and it is always written in OWL 2 in double quotes, and *datatype* is the datatype of the attribute *Value*.

*In case of multiplicity constraints*: The OCL constraint `self.A > size()=Value` (see Fig. 24), in which $A$ is the name of an association, " > " is an infix operator and *Value* is a positive integer, is translated in OWL 2 as

```
EquivalentClasses (State
  ObjectExactCardinality(Value A))
```

*In case of boolean operators*: The OCL constraint self.Att = Valueandself.A − ≥ size()=Value    shown    in Fig. 25 is translated in OWL 2 as

```
EquivalentClasses (State
  ObjectIntersectionOf(
  DataHasValue(Att }Value}^^datatype)
  ObjectExactCardinality(Value A)))
```

## 9. Application: consistency of multiple merged models

The process of merging multiple models using OWL 2 is already discussed in the previous sections. In this section we discuss how to check the consistency of merged models using OWL 2 reasoners. To explain the consistency of merged models, we assume that there is a nonempty set $\Delta^{\mathcal{I}}$ called the object domain containing all the possible objects in our domain. We propose that a merged model depicting a class diagram is interpreted as a number of subsets of $\Delta^{\mathcal{I}}$ representing each class in the merged model and as a number of conditions that need to be satisfied by these sets. The merged model is consistent, if each class in a merged model can be instantiated i.e, if $C$ is a class in a merged model and $C \subseteq \Delta^{\mathcal{I}}$ then $C \not\equiv \bot^{\mathcal{I}}$ must hold.

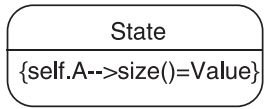

Fig. 24. OCL state invariant – multiplicity constraints.

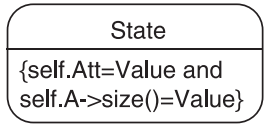

Fig. 25. OCL state invariant – boolean operators.

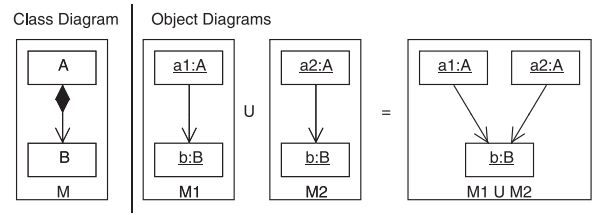

Fig. 26. The invalid merge of two valid UML models.

Moreover, in order to check the model conformance (that an object diagram conforms to the class diagram), we want to know, if each object $o$ in an object diagram is a proper instance of a class $C$ in a class diagram i.e, $o \in C$. Moreover, we want to investigate that each link in an object diagram is an instance of an association depicted in a class diagram. Also, objects and links must preserve the constraints applied in a class diagram such as uniqueness, multiplicity, source, target, ordering and composition constraints of their classes and associations.

### 9.1. Validation of merged models

The consistent UML models representing different versions of a UML model may have inconsistencies when viewed together or merged. For example UML models $M1$ and $M2$ shown in Fig. 26 are valid and consistent models, but when they are merged, the merged model $M1 \cup M2$ is inconsistent, because it is violating the exclusive ownership constraint of composition. Exclusive ownership means that an object can have only one owner.

In order to detect the inconsistencies occurred by the merging of models, we first translate the models into the OWL 2 ontology, and then use an OWL 2 reasoner to detect the inconsistencies in the translated ontology. To demonstrate our validation approach we have translated the models $M$, $M1$ and $M2$ shown in Fig. 26 into an ontology. The generated ontology is shown in Fig. 27.

### 9.2. Reasoning

In order to detect the inconsistencies in an ontology (Fig. 27) of a merged model (Fig. 26), the OWL 2 ontology of the merged model is passed to the OWL 2 reasoner. Since the ontology is in OWL2fs format, we can use any reasoner which supports this format. In our example we have used Pellet [20] version 2.3.0. The reasoner processes the ontology and generates a validation report.

The validation report indicates the contradictions and inconsistencies in the ontology caused by the model merge. The validation report of the ontology (shown in Fig. 27) of models (Fig. 26) is as follows:

```
D:\pellet-2.3.0>pellet consistency -l
OWLAPI D:\METest.owl2fs
```

....

```
//Class Diagram M

Declaration(Class(A))

Declaration(Class(B))

Declaration(ObjectProperty( association_A_B ))

ObjectPropertyDomain(association_A_B A)

ObjectPropertyRange(association_A_B B)

SubObjectPropertyOf(association_A_B owns )

//Object diagram M1

SubClassOf( ObjectOneOf( A ) a1 )

SubClassOf( ObjectOneOf( B ) b )

ObjectPropertyAssertion( association_A_B a1 b)

//Object diagram M2

SubClassOf( ObjectOneOf( A ) a2 )

SubClassOf( ObjectOneOf( B ) b )

ObjectPropertyAssertion( association_A_B a2 b)

DifferentIndividuals( a1 a2 b )
```

.....

Fig. 27. The OWL 2 ontology of the models $M$, $M1$ and $M2$ (Fig. 26).

```
Consistent: No
Reason: Individual file:D:\METest.owl2fs#b
has more than one value for the functional
property inv(file:D:/METest.owl2fs#owns)
```

The validation report clearly indicates that the violation of the mutually exclusive ownership constraint of composition is that the object $b$ has more than one value for the inverse functional property *owns*. In our OWL 2 translations of a composition the inverse-functional object-property *owns* represents the exclusive ownership constraint of the composition. The details about the translation of composition constraints have been discussed in the previous sections.

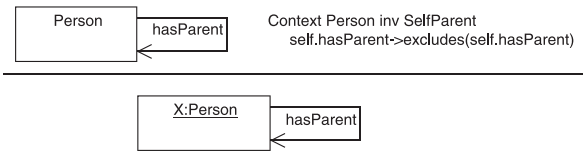

Context Person inv SelfParent
self.hasParent->excludes(self.hasParent)

**Fig. 28.** Top: a UML class diagram with OCL constraints. Bottom: the object diagram of a class diagram depicted on top.

## 10. Application: conformance of objects against OCL constraints

In order to check the conformance of an object diagram against a class diagram with OCL constraints, we need to first translate these diagrams and OCL constraints into OWL 2 by following the translations discussed in the previous sections. The next step is to validate the OWL 2 translations by using an OWL 2 reasoner. As an example we have a class diagram shown in Fig. 28 consisting of a class named *Person*, and the association named *hasParent*. Along with the class diagram there is an OCL constraint name SelfParent which has a context class *Person*. This OCL constraint is applying a restriction on the objects of a class person that they cannot link to themselves by using an instance/link of an association *hasParent*. At the bottom of Fig. 28 we have an object diagram depicting an object *X* of the class *Person*. This object is linked with itself by using a link of the association *hasParent*. In order to determine that the object diagram conforms to the class diagram, we first translate these diagrams and OCL constraint into OWL 2.

### 10.1. OCL to OWL 2

The translation of a class diagram and an object diagram along with OCL constraints (Fig. 28) into OWL 2 is shown in Fig. 29.

### 10.2. Reasoning

After translating the class and object diagram with OCL constraint we pass the OWL 2 ontology to the OWL 2 reasoner that generates the validation report. The validation report of the ontology of the diagrams and constraint shown in Fig. 28 is as follows:

```
D:\pellet-2.3.0>pellet consistency -1
OWLAPID:\OCL.OWL2FS
Consistent: No
Reason: Irreflexive property nullhasParent
```

The above validation report clearly shows the violation of the OCL constraint that an object cannot link with its own self by using a link of the association *hasParent* i.e. *Irreflexive*(*hasParent*). Due to the violation of the OCL constraint the reasoner finds the ontology inconsistent.

```
//Class Diagram

Declaration(Class(Person))

Declaration(ObjectProperty( hasParent ))

ObjectPropertyDomain(hasParent Person)

ObjectPropertyRange(hasParent Person)

//OCL Constraint

IrreflexiveObjectProperty(hasParent)

//Object Diagram

SubClassOf( ObjectOneOf( X ) Person )

ObjectPropertyAssertion( hasParent X X)
```

**Fig. 29.** The OWL 2 translation of models shown in Fig. 28.

## 11. Application: consistency of class diagrams and statechart diagrams

In this section we present an overview of our approach that we demonstrate with a running example. Our example system is a Content Management System (CMS). In this system, authors post new articles to be published after being reviewed by a reviewer. A reviewer can accept,reject or advise a revision of the article. Only an accepted article can be published. An article can be withdrawn if it is under review. However, a published article cannot be withdrawn. The structure of this system is described as a UML class diagram (Fig. 30), while its behavior is described using a UML statechart diagram (Fig. 31).

### 11.1. Class diagram representing structure of CMS

The class diagram provides the main classes involved in the system under development and their associations with each other. It exposes the attributes of each class and operations that can be invoked on them.

The class diagram shown in Fig. 30 shows the syntactic view of CMS. It consists of 5 classes, namely, *Article*, *Review*, *Withdraw*, *PublicationRecord* and an enumeration class *DecisionType*. An article is associated to *Review*, *Withdraw* and *PublicationRecord* classes via associations *review*, *withdraw* and *publicationRecord*, respectively. *Review* class is further associated to an enumeration class *DecisionType* with literals accept, reject and revise. An instance of the article class can be submitted, withdrawn, published or revisioned via *Submit*, *Withdraw*, *Publish*and *Revisioned* operations, respectively. An *accept*, *reject* and *revise* operation can be called on an instance of *Review* class.

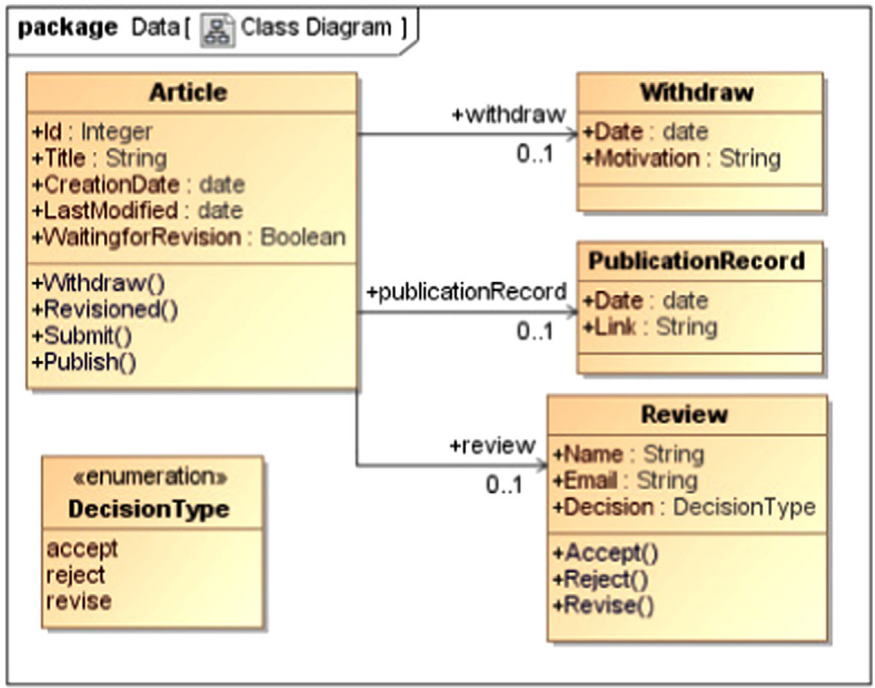

**Fig. 30.** The static view of content management system.

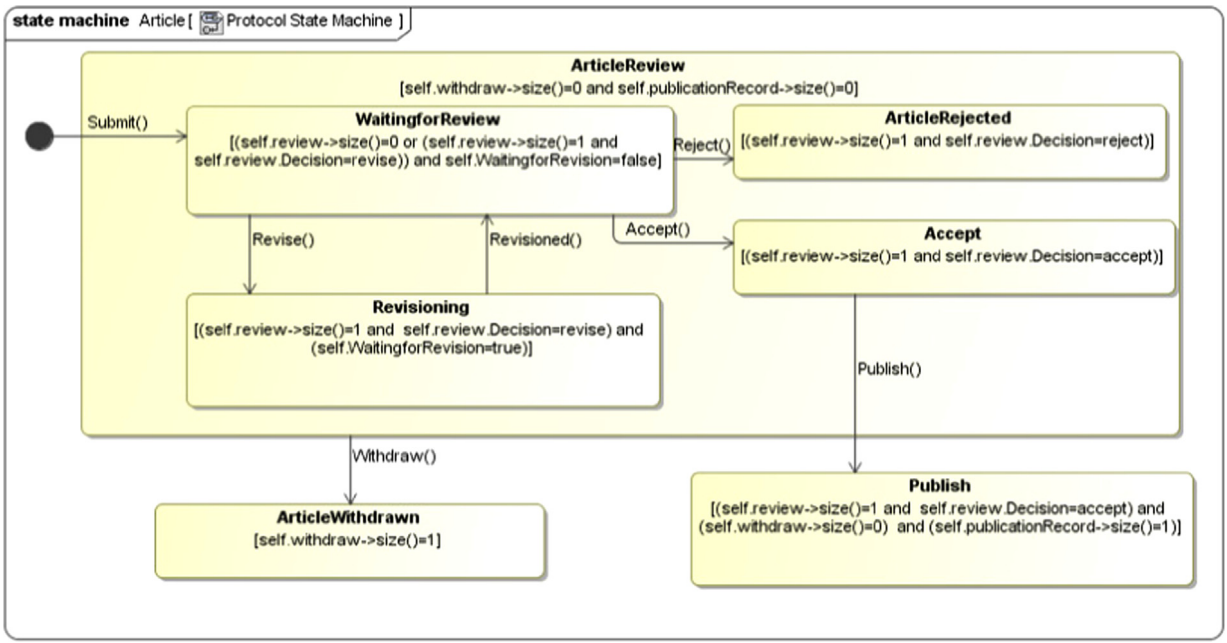

**Fig. 31.** The behavioral view of the class *Article* of the class diagram shown in Fig. 30.

### 11.2. Statechart diagram representing behavior of CMS

A statechart diagram defines behavior of a class in terms of states that an instance of a class takes during its lifecycle and the transitions between them. Each transition from a source to a target state is triggered by a function call.

The statechart diagram shown in Fig. 31 defines the behavioral view of the class *Article* of the class diagram shown in Fig. 30 in terms of states. It consists of one composite state *ArticleReview* and two simple states *Publish* and *ArticleWithdraw*. The *ArticleReview* composite state consists of four simple states namely, *WaitingforReview*, *Revisioning*, *ArticleRejected* and *Accept*. When the *submit*() method is called on an object of the class *Article*, the statechart diagram is initiated and the object enters into the state *WaitingforReview*, a substate of *ArticleReview*. The method calls *accept*(), *reject*() and *revise*() take the object to the *Accept*, *ArticleRejected* and *Revisioning* states respectively. When the author of the article is revisioning the article, the object of the class *Article* is in the *Revisioning* state. When the author revises the article, he invokes the *Revisioned*() method of the *Article* class and the object again comes into the *WaitingforReview* state. The *publish*() method can be invoked from the *Accept* state and the object switches to the *Publish* state. An article can be withdrawn by invoking the method *withdraw*() whenever the state *ArticleReview* is active, but the *withdraw*() method cannot be invoked if the object of the class *Article* is in the *Publish* state.

### 11.3. State invariants

Each state in a statechart is annotated with a state invariant. The state invariant is a boolean expression that links classes of a class diagram to the states of a statechart diagram. We say that an object of a class is in a certain state if the state invariant of that state is true. We express the state invariant of each state by using OCL and annotate the behavioral diagram of our example with state invariants in Fig. 31. The details about the OCL constructs used in our approach have been discussed in Section 8.5.

### 11.4. Invalid state invariant

We consider the state invariants which let the statechart diagram behave against the UML superstructure specifications for statechart diagrams [2] as inconsistent state invariants, and they may cause the whole system to become unsatisfiable or inconsistent. The examples of inconsistent state invariants are as follows:

*Inconsistent State Invariant Example* 1: According to the UML superstructure specification, invariants of nonorthogonal states must be mutually exclusive [2, p. 564], for example in the statechart diagram shown in Fig. 31, the article cannot be in the state *ArticleRejected* if at the same time this article is in the state *Accept*. If we introduce an error by changing the invariant value of the state *ArticleRejected* to

```
self.review->size()=1 and
  self.review.Decision=accept
```

this means that an article can be rejected and accepted at the same time. The introduced error allows an object of the class *Article* to belong to two non-orthogonal states i.e. *Accept* and *ArticleRejected*, which is the violation of the UML superstructure specification of the statechart diagram, and as a consequence the invariants of states *ActicleRejected* and *Accept* become inconsistent.

*Inconsistent State Invariant Example* 2: According to the UML superstructure specification, whenever a state is active, all its superstates are active [2, p. 565]. This means that all the invariants of an active state and its superstates directly or transitively are true. For example, in the statechart diagram of Fig. 31, if the state *Accept* is active then its superstate *ArticleReview* should be also active. If we introduce an error by adding the condition self.withdraw->size()=1 in the invariant of the state *Accept*, this means that a withdrawn article can also be accepted. The introduced error causes the contradiction between the invariants of the state *Accept* and its superstate *ArticleReview*, and violates the UML superstructure specification of the statechart diagram. Consequently it makes the invariant of the states *Accept*, *ArticleReview* and *ArticleWithdrawn* inconsistent.

In the next section we discuss how we can carry out the analysis of these kind of models using OWL 2 reasoning tools.

### 11.5. Consistency analysis

In this section we define the problem of determining the consistency of UML models containing class and statechart diagrams as follows. Our view of model consistency is inspired by the work of Broy et al. [63]. This work considers the semantics of a UML diagram as their denotation in terms of the so-called system model and defines a set of diagrams as consistent when the intersection of their semantic interpretation is nonempty.

In our work, we assume that there is a nonempty set $\Delta$ called the object domain containing all the possible objects in our domain. We propose that a UML model depicting a number of class and statechart diagrams is interpreted as a number of subsets of $\Delta$ representing each class and each state in the model and as a number of conditions that need to be satisfied by these sets.

A UML class is represented by a set $C$, such that $C \subseteq \Delta$. An object o belongs to a UML class $C$ iff $o \in C$. We also represent each state $S$ in a statechart as a subset of our domain $S \subseteq \Delta$. In this interpretation, the state set $S$ represents all the objects in the domain that have such state active, i.e., object o is in UML state $S$ iff $o \in S$.

Other elements that can appear in a UML model such as generalization of classes, association of classes, state hierarchy and state invariants are interpreted as additional conditions over the sets representing classes and states. For example class specialization is interpreted as a condition stating that the set representing a subclass is a subset of the set representing its superclass. These conditions has been described in detail in the previous sections.

In this interpretation, the problem of a UML model consistency is then reduced to the problem of satisfiability of the conjunction of all the conditions derived from the model. If such conditions cannot be satisfied, then a UML model will describe one or more UML classes that cannot be instantiated into objects or objects that cannot ever enter a UML state in a statechart. This can be considered a design error, except in the rare occasion that a designer is purposely describing a system that cannot be realized. To analyze the UML models and discover possible inconsistencies we will use the services of an OWL 2 reasoning tool, as described in the rest of this section.

### 11.6. Reasoning

A number of UML class diagrams, statechart diagrams and state invariants are taken as an input. All the inputs are translated to OWL 2 DL, and then analyzed by a reasoner. The reasoner provides a report of unsatisfiable and satisfiable concepts. Unsatisfiable concepts will reveal UML classes that cannot be instantiated or UML states that cannot be entered.

### 11.7. Consistency analysis using an OWL 2 reasoner

We have defined earlier the satisfiability of UML models in Section 11.5. The consistency analysis of UML models is reduced to the satisfiability of the conjunction of all conditions derived from a model. In order to determine the satisfiability of the conditions represented in UML models, we first translate the UML models into an OWL 2 ontology, then use an OWL 2 reasoner to analyze the satisfiability of translated concepts.

To translate UML models into OWL 2 ontology, we have implemented the translations of class diagrams, statechart diagrams and state invariants discussed in the previous sections, in an automatic model to text translation tool. The implemented translation tool allows us to automatically translate class diagrams, statechart diagrams and state invariants into OWL 2 DL. The translator reads class diagrams, statechart diagrams and OCL state invariants from an input model serialized using the XMI format. The XMI is generated by using a modeling tool. We used Magicdraw to create the example design used in this section. The output of the translation tool is an ontology file ready to be processed by an OWL 2 reasoner.

As an example, we have translated the class diagram, statechart diagram and OCL state invariants shown in Figs. 30 and 31, into OWL 2 DL ontology using the implemented translation tool. An excerpt of the output ontology generated by the translation tool is shown in Fig. 32.

### 11.8. Reasoning

After translating the class diagram, statechart diagram and state invariants into an OWL 2 ontology by using the implemented translation tool, we process the ontology by using an OWL 2 reasoner. The OWL 2 reasoner combines all the facts presented as axioms in the ontology and infers logical consequences from them. When we give the generated ontology to the reasoner, it generates a satisfiability report indicating which concepts are satisfiable and which not. If the ontology has one or more unsatisfiable concept, this means that the instance of any unsatisfiable concept will make the whole ontology inconsistent. Consequently, an instance of the class describing an unsatisfiable concept in a class diagram will not exist, or objects will not enter into a state describing an unsatisfiable condition. However, in case of a consistent ontology an object of a class can be created and the object may also enter in a state.

In order to analyze the satisfiability of the inconsistent invariants listed in Section 11.4, the ontology of an example model with inconsistent invariants is validated by using Pellet. The satisfiability report of the ontology of UML models with inconsistent state invariants is shown in Fig. 33.

As explained in Section 8.4, a state invariant characterizes the state [2, pp. 559–560]. Therefore, the presence of unsatisfiable states in the satisfiability report indicates the existence of inconsistent state invariants in identified states.

## 12. Performance analysis

In order to determine the performance of the translation and reasoning tools, we conducted an experiment using UML class diagrams and statechart diagrams with and without invariants consisting of 10–2000 model elements. We use a desktop computer with an Intel Core 2 Duo E8500 processor running at 3.16 GHz with 2 GB of RAM. The performance tests are conducted for both consistent and mutated models containing inconsistencies introduced by us. For each test,we measure the time required to translate a model from UML to OWL 2 and the time required by the OWL 2 reasoners Pellet and HermiT to analyze the models. The results are shown in Table 4, and in Fig. 34.

The time complexity of OWL 2 DL with respect to the reasoning problems of the ontology consistency and instance analyzing is NEXPTIME complete [30]. However, the graph (Fig. 34) of the performance test shows that the time required to reason about models only grows linearly. This is due to the fact that in our approach we analyze the consistency of class and statechart diagrams without individuals.

## 13. Conclusion

In this paper we have presented an approach to automatically analyze the consistency and satisfiability of UML models containing multiple class, object and statechart diagrams using OWL 2 reasoners. We showed how to translate the UML models into OWL 2, and how to use OWL 2 reasoners to analyze the translated UML models.

The proposed UML to OWL 2 translation has been implemented in a tool. Since the translation tool accepts UML 2 version 3.0.0 as an input, it is possible to integrate the translation tool with any standard compliant UML modeling tool that follows this UML standard.

We have validated this approach using different scenarios. Each scenario has been validated by using both valid and mutated models comprising of up to 2000 model elements, and the detection of mutants during the validation process clearly indicated the usefulness of the proposed approach. At the same time, the performance of this approach was also analyzed using these models. The translation and the reasoning time of all models was less than 4.5 s in all cases. This shows that the proposed approach can process relatively large UML models in a few seconds.

A future improvement in the current work would be to enhance the way problems in ontologies are reported. In few cases, the relationship between UML concepts and OWL 2 axioms is not obvious and it is not possible to immediately point out the cause of the problem based on these violations without a detailed inspection of the validation report and the problematic models. It would

```
// Class Diagram into OWL 2 DL
Declaration(Class(Article))
Declaration(Class(Review))
Declaration(Class(Withdraw))
Declaration(Class(PublicationRecord))
...
DisjointClasses( Article Review ...)
Declaration(ObjectProperty(review))
ObjectPropertyDomain( review Article)
ObjectPropertyRange( review Review )
...
SubClassOf( Article
ObjectMaxCardinality( 1 review ))
....
Declaration(
DataProperty( WaitingforRevision))
SubClassOf(Article
DataExactCardinality(1
        WaitingforRevision))
...
DataPropertyDomain(
WaitingforRevision Article )
..
DataPropertyRange(
WaitingforRevision xsd:boolean )
```

```
//Statechart diagram into OWL 2 DL

Declaration(Class(ArticleReview))
Declaration(Class(ArticleWithdraw))
Declaration(Class(Publish))
SubClassOf( ArticleReview Article )
SubClassOf( ArticleWithdraw Article )
SubClassOf( Publish Article ))
...
DisjointClasses( ArticleReview
ArticleWithdraw Publish )
Declaration(Class(WaitingforReview))
SubClassOf( WaitingforReview ArticleReview )
...
//Invariant of state Publish Start
EquivalentClasses (Publish
ObjectIntersectionOf(
ObjectIntersectionOf(
ObjectExactCardinality(1 review)
DataHasValue(Decision "accept"^^xsd:string ))
ObjectIntersectionOf ( ObjectExactCardinality
(0 withdraw) ObjectExactCardinality
(1 publicationRecord)) ) )
//Invariant of state Publish End
```

**Fig. 32.** Excerpt of the output ontology generated by the translation tool.

```
Found 4 unsatisfiable concept(s):
a:Accept
a:ArticleRejected
a:ArticleReview
a:ArticleWithdrawn
```

**Fig. 33.** The satisfiability report of the ontology shown in Fig. 32 generated by the OWL 2 reasoner Pellet.

**Table 4**
Time taken by the translation tool and reasoning engines to process UML models.

| Model elements | 10 (s) | 100 (s) | 500 (s) | 1000 (s) | 1500 (s) | 2000 (s) |
|---|---|---|---|---|---|---|
| **Translation time** | 0.08 | 0.11 | 0.19 | 0.30 | 0.44 | 0.53 |
| **Pellet** | | | | | | |
| Valid | 2.2 | 2.3 | 2.6 | 3.2 | 3.6 | 3.8 |
| Mutated | 2.2 | 2.4 | 2.7 | 3.2 | 3.6 | 3.9 |
| **HermiT** | | | | | | |
| Valid | 0.6 | 0.7 | 1.2 | 1.7 | 2.2 | 2.6 |
| Mutated | 0.7 | 0.7 | 1.3 | 1.8 | 2.3 | 2.6 |

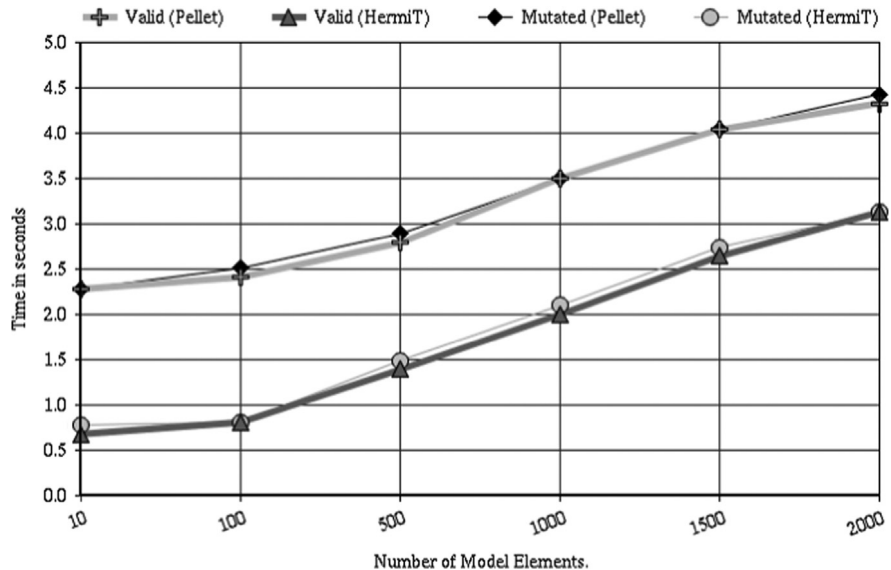

**Fig. 34.** Total time (translation time+reasoning time) to process valid and mutated models.

greatly add to the usefulness of the method to have some sort of automated discovery of the cause of violations.